\documentclass[10pt,twocolumn,letterpaper]{article}

\usepackage[pagenumbers]{cvpr} % To force page numbers, e.g. for an arXiv version

\usepackage{graphicx}
\usepackage{amsmath}
\usepackage{amssymb}
\usepackage{booktabs}

\usepackage{multirow}
\usepackage{tabularx,verbatim}
\usepackage{xspace}
\usepackage{algorithm}
\usepackage{algorithmic}
\usepackage{setspace}
\usepackage{comment}
\usepackage{bbm}
\usepackage{enumitem}
\usepackage{colortbl}
\usepackage{color, xcolor} % colors
\usepackage[T1]{fontenc}
\usepackage{bbm}
\usepackage{listings}

\usepackage{pifont}% http://ctan.org/pkg/pifont
\newcommand{\cmark}{\ding{51}}%
\newcommand{\xmark}{\ding{55}}%

\definecolor{remark}{rgb}{1,.5,0} 
\definecolor{citecolor}{rgb}{0,0.443,0.737} 
\definecolor{linkcolor}{rgb}{0.956,0.298,0.235} 
\definecolor{cyan}{rgb}{0.831,0.901,0.945}

\usepackage{amsmath}

\usepackage[pagebackref,breaklinks,colorlinks,bookmarks=false]{hyperref}
\colorlet{dark-blue}{blue!70!black}
\colorlet{dark-green}{green!80!black}
\colorlet{dark-red}{red!80!black}
\definecolor{mypink}{RGB}{219, 48, 122}
\hypersetup{
    colorlinks=true,%
    citecolor=citecolor,%
    filecolor=citecolor,%
    linkcolor=linkcolor,%
    urlcolor=magenta
}
\usepackage{url}

\usepackage[capitalize]{cleveref}
\crefname{section}{Sec.}{Secs.}
\Crefname{section}{Section}{Sections}
\Crefname{table}{Table}{Tables}
\crefname{table}{Tab.}{Tabs.}

\def\cvprPaperID{383} % *** Enter the CVPR Paper ID here
\def\confName{CVPR}
\def\confYear{2023}

\begin{document}

\title{Synthetic-to-Real Domain Generalized Semantic Segmentation \\ for 3D Indoor Point Clouds}

\author{%
  Yuyang Zhao$^{\textcolor{mypink}{1}}$ \quad Na Zhao$^{\textcolor{mypink}{2}}$ \quad Gim Hee Lee$^{\textcolor{mypink}{1}}$ \\
  \small{$^{\textcolor{mypink}{1}}$ Department of Computer Science, National University of Singapore}  \\
  \small{$^{\textcolor{mypink}{2}}$ Information Systems Technology and Design Pillar, Singapore University of Technology and Design}
}

\maketitle

%%%%%%%%% ABSTRACT
\begin{abstract}

Semantic segmentation in 3D indoor scenes has achieved remarkable performance under the supervision of large-scale annotated data. However, previous works rely on the assumption that the training and testing data are of the same distribution, which may suffer from performance degradation when evaluated on the out-of-distribution scenes. 
To alleviate the annotation cost and the performance degradation, this paper introduces the synthetic-to-real domain generalization setting to this task.
Specifically, the domain gap between synthetic and real-world point cloud data mainly lies in the different layouts and point patterns. 
To address these problems, we first propose a clustering instance mix (CINMix) augmentation technique to diversify the layouts of the source data.
In addition, we augment the point patterns of the source data and introduce non-parametric multi-prototypes to ameliorate the intra-class variance enlarged by the augmented point patterns. The multi-prototypes can model the intra-class variance and rectify the global classifier in both training and inference stages.
Experiments on the synthetic-to-real benchmark demonstrate that both CINMix and multi-prototypes can narrow the distribution gap and thus improve the generalization ability on real-world datasets.

\end{abstract}

%%%%%%%%% BODY TEXT
\section{Introduction}
\label{sec:intro}

3D point cloud semantic segmentation is one of the most crucial and fundamental tasks in point cloud based scene understanding, and the semantic segmentation of indoor categories has great potential for real-world applications, such as robotics~\cite{ambrucs2017automatic} and building information management~\cite{yin2021automated}.
With the development of deep neural networks and abundant annotated data, modern models for 3D semantic segmentation~\cite{qi2017pointnet++,wang2019dynamic,qian2022pointnext} have achieved remarkable success in several benchmark datasets~\cite{scannet,s3dis} under the fully-supervised setting. 
However, modern deep neural networks generally suffer from performance degradation when evaluating on the data with different distributions from the training ones. 
Specifically, the layout of indoor scenes can be varied in different countries or even different people, and the point patterns of the point clouds can be diverse due to the change of 3D data capture devices and procedures. 
As a result, the performance of modern 3D semantic segmentation methods is not guaranteed with the existence of domain shift (\eg, different layout and point patterns) between the training and testing data, which are inevitable in real-world applications. Moreover, the annotation cost for the 3D semantic segmentation is enormously expensive, which hinders the acquisition of a highly generalizable model that requires diverse annotated data for training.
%.

\begin{figure}[t]
    \centering
    \includegraphics[width=.99\linewidth]{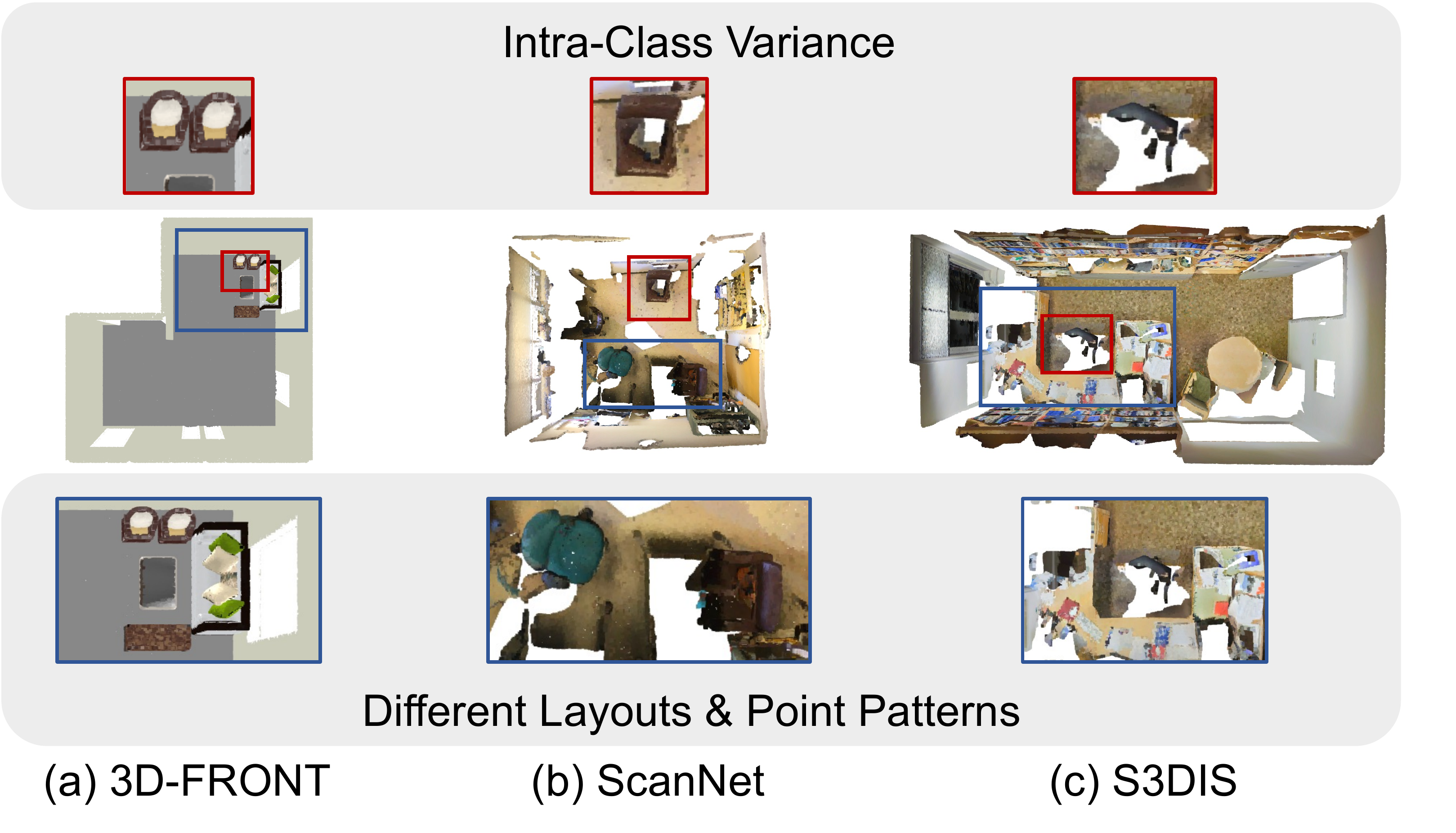}
    \caption{Distribution gap between the synthetic source ((a) 3D-FRONT~\cite{3dfront}) and real-world target ((b) ScanNet~\cite{scannet} and (c) S3DIS~\cite{s3dis}) data. The distribution gap mainly lies in different layouts and point patterns, which exacerbate the intra-class variance.}
    \label{fig:intro}
\end{figure}

To mitigate the performance degradation and heavy annotation cost of 3D indoor point clouds, we first introduce the synthetic-to-real domain generalization for 3D indoor point cloud semantic segmentation (DG-Indoor-Seg), which only leverages one synthetic dataset to train a robust model that can generalize well to various testing environments in the real-world. 
Domain generalized semantic segmentation has been investigated in 2D driving scenes~\cite{robustnet,zhao2022shade,zhong2022adversarial}, where the 2D domain shift is mainly caused by various weather, time, and conditions in different streets.
Unlike the 2D outdoor scenarios, the domain shift between synthetic and real-world data in 3D indoor scenarios is more complex, including the layouts and point pattern differences. As shown in Fig.~\ref{fig:intro}, the synthetic 3D scene from 3D-FRONT dataset~\cite{3dfront} has a clean and ordered layout since the scene layout in this dataset is designed by some experts, and the point cloud of the synthetic scene is dense and complete. In contrast, the real-world 3D scenes from ScanNet~\cite{scannet} and S3DIS~\cite{s3dis} datasets are with cluttered layouts and complex object-furniture relationships. Additionally, compared to the synthetic point cloud, the point clouds of these real scenes are relatively sparse and noisy with severe occlusions. 
Based on the observations of the two types of domain shift, we address the DG-Indoor-Seg by diversifying the synthetic layouts and 
investigating the shift of feature representations caused by point pattern augmentation.
2D domain generalization methods~\cite{tranheden2021dacs,pan2022ml} utilize mixing patches~\cite{yun2019cutmix} or classes~\cite{olsson2021classmix} to diversify the training data with additional context, which can improve the generalization ability in 2D driving scenes.
Compared with regular 2D images, 3D point clouds are spatially irregular with varying sizes. Thus, a direct implementation of these 2D mixing techniques may result in unsatisfactory augmentations. Recently, a few works~\cite{nekrasov2021mix3d,ding2022doda} study 3D mixing techniques by mixing the whole scenes or cuboids. Despite the performance improvements on in-distribution testing data~\cite{nekrasov2021mix3d} or adaptive target data~\cite{ding2022doda}, the naive mixing of scenes or cuboids leads to unrealistic augmentations, which cannot benefit out-of-distribution data and might deteriorate the performance of domain generalization. Instead of the mixing two scenes/cuboids without any constraint, we propose a Clustering INstance Mix (CINMix) technique, which mixes the object instances in one scene with another scene under several rational geometry constraints, to generate diverse and realistic scenes.
Specifically, due to the unavailability of instance-level labels in semantic segmentation, we apply density-based clustering on each class to roughly separate instances within one class and treat the resultant clusters as instances. Subsequently, we sample multiple instances and place them on a free location of the floor in an arbitrary scene. Equipped with CINMix, we can generate more source scenes with cluttered and diverse layouts.

As mentioned earlier, another severe domain shift for synthetic-to-real DG-Indoor-Seg is in form of different point patterns between synthetic and real domains. 
To narrow the domain gap, we follow \cite{ding2022doda} to add noise and occlusion to the training data. 
As can be seen from the real scenes in Fig.~\ref{fig:intro}, occlusion and noise would change the geometrical shape of objects.
Since 3D semantic segmentation extracts features based on the geometrical information, the augmentation on the point patterns leads to larger intra-class variance of the source domain. With the enlarged intra-class variance, the original global classifier in semantic segmentation that learns a single weight vector for each class may become insufficient. To compensate the global classifier and model the large intra-class variance, we introduce non-parametric multi-prototypes that represent each class with multiple prototypes. The multi-prototypes encourage the model to discover diverse and discriminative patterns within each class and optimize the mining of the instance-specific information.
Specifically, we use clustering to obtain several prototypes for each class as the initialization, and then update them by moving average. 
The update process is formulated as the optimal transport problem to uniformly update all the prototypes.
During training, the multi-prototypes are used as the non-parametric classifier containing rich intra-class information. 
During inference, the well-learned multi-prototypes can also be used to rectify the model predictions from the global classifier, thanks to their encoded instance-specific information. 

Our main contributions can be summarized as follows:
\begin{itemize}
    \item We propose a practical yet challenging setting of domain generalized 3D indoor point cloud semantic segmentation (DG-Indoor-Seg), %
    which trains a robust model on a synthetic source dataset but the model is able to segment point clouds from other real-world datasets. 
    
    \item We propose a novel data augmentation technique, \ie, clustering instance mix (CINMix), for synthetic-to-real DG-Indoor-Seg. CINMix can generate more realistic scenes with complex and cluttered layouts for the source data to narrow the layout domain shift. 
    
    \item We propose a non-parametric multi-prototypes based classifier to deal with the intra-class variance exacerbated by the point pattern augmentation. The multi-prototypes can be used in both training and inference stages to rectify the model.
    
\end{itemize}

\section{Related Work}
\begin{figure*}[t]
    \centering
    \includegraphics[width=.99\linewidth]{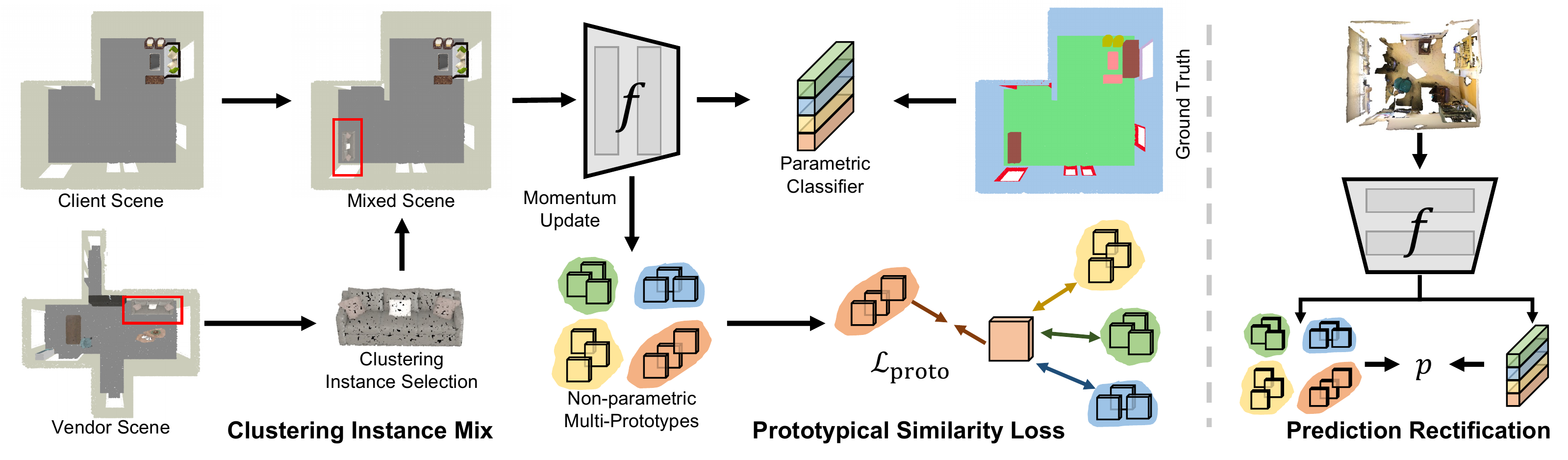}
    \vspace{-.1in}
    \caption{{The overall framework of our model for DG-Indoor-Seg}. During training, CINMix is first leveraged to generate diverse source samples, and then the model is trained with these samples via the standard cross entropy loss and a prototypical similarity loss based on the momentum-updated multi-prototypes. During inference, the multi-prototypes are used to rectify the model prediction.}
    \label{fig:framework}
\end{figure*}

\noindent\textbf{3D Indoor Point Cloud Semantic Segmentation.}
In the deep learning era, point-based~\cite{qi2017pointnet++,qian2022pointnext,wang2019dynamic} and voxel-based~\cite{choy20194d,riegler2017octnet,sparseconv} deep neural networks have achieved significant performance under the supervision of fully annotated data. Considering the difficulties in annotating point clouds, semi-supervised~\cite{cheng2021sspc,jiang2021guided} and few-shot~\cite{zhao2021few,zhao2022crossmodal} settings have drawn more attention in recent years. 
More recently, Ding \etal~\cite{ding2022doda} investigate the domain adaptive semantic segmentation in 3D indoor scenes, where labeled source data and unlabeled target data are used to learn a adaptive model that can perform well on the target testing data.
Different from previous works, in this paper, we first introduce the practical yet challenging domain generalization setting to 3D indoor point cloud semantic segmentation.

\noindent\textbf{Domain Generalization.}
To tackle the performance degradation in the different domains, domain generalization~\cite{li2017learning,zhao2021learning,ibn,zhao2022sfocda,zhao2022shade,zhong2022adversarial} is widely explored in the community, which learns a robust model with one or multiple source domain(s), aiming to perform well on unseen domains. In recent years, domain generalized semantic segmentation~\cite{zhao2022sfocda,zhao2022shade,robustnet,DRPC} in 2D driving scenes has been studied, where the domain shift lies in the changeable environments and styles. Briefly, one mainstream of these works focuses on designing style-invariant modules, such as whitening~\cite{robustnet} and instance normalization~\cite{ibn}. Another mainstream engages in diversifying the source samples with style augmentation~\cite{zhao2022shade,zhong2022adversarial} and image randomization~\cite{DRPC,FSDR}.
Different from previous works on 2D driving scenes, we study domain generalized semantic segmentation on 3D indoor point clouds, where the domain shift lies in the point patterns and layouts instead of image styles.

\noindent\textbf{Data Augmentation for Semantic Segmentation.}
In addition to standard transformations, such as geometry transformation and color jitter, many advanced data augmentation techniques can improve both in-distribution and out-of-distribution performance. CutMix~\cite{yun2019cutmix} crops a patch from one sample and mixes it with another sample, which can improve the performance of both semantic segmentation and image classification. ClassMix~\cite{olsson2021classmix} is proposed for semi-supervised semantic segmentation, which takes pixels of several classes in one image to another one. 
As for 3D point clouds, Mix3D~\cite{nekrasov2021mix3d} directly concatenates two samples to train the semantic segmentation model. Cuboid Mixing~\cite{ding2022doda} is designed for domain adaptive 3D point cloud semantic segmentation, which splits each scene into several cuboids and mixes the cuboids from source and target domains.
In this paper, we propose the CINMix to
generate diverse and realistic layouts within the source domain to overcome the domain shift regarding layout change.

\noindent\textbf{Non-parametric Prototypes.}
Non-parametric prototypes are mainly investigated in few-shot learning~\cite{snell2017prototypical,dong2018few}, where they use features in the support set as the prototypes for matching and classification. 
Later, to model the complex data distribution in one episode, Zhao \etal~\cite{zhao2021few} leverage multiple prototypes for each class to classify and propagate labels. The concept of non-parametric prototypes is also applied to the fully-supervised semantic segmentation. Zhou \etal~\cite{zhou2022rethinking} replace the parametric classifier with non-parametric prototypes to gain improvement in various segmentation models.
Unlike previous works, the non-parametric multi-prototypes in this paper is used to model the intra-class variance and rectify the parametric classifier in both training and inference stages.

\section{Methodology}
\subsection{Overview}
% setting and framework
In this paper, we focus on domain generalized semantic segmentation in 3D indoor scenes, where one synthetic dataset $D_{S}=\{ \mathcal{X}_S, \mathcal{Y}_S \}$ is given as the source training data and the model is expected to perform on $N_T$ unseen target domains $D_{T}=\{ \mathcal{X}_T, \mathcal{Y}_T \}_i^{N_T}$. The source and target domains share the same label space ($N_C$ categories), but have different distributions. As illustrated in the introduction, the domain gap mainly lies in the different layouts and point patterns. Consequently, we propose the Clustering INstance Mix (CINMix) to diversify source layouts and non-parametric multi-prototypes to address the intra-class variance exacerbated by point pattern augmentations. 
The overall framework of our proposed method is shown in Fig.~\ref{fig:framework}. Specifically, 
{we randomly select the point clouds of two scenes during training, and consider one point cloud scene as vendor scene (to sample instances from it) while treating the other one as client scene (to insert the sampled instances into it).}
The thing classes of the vendor scene are clustered into several groups, and then the available groups are mixed to the client point cloud scene subjecting to the geometry constraints. 
After the CINMix step, the mixed sample is augmented with virtual scan simulation~\cite{ding2022doda} and standard geometry augmentation techniques. 
Then it is fed into the model that consists of a feature encoder and a parametric classifier to generate its prediction, which is supervised by the ground truth via the standard cross entropy loss. 
In addition, we utilize non-parametric multi-prototypes to compute a prototypical similarity loss as the additional supervision. The non-parametric multi-prototypes are used to model the intra-class variance and is updated by moving average.
During inference, the multi-prototypes are used to rectify the prediction from the parametric classifier. With the help of CINMix and multi-prototypes, our model can deal with various unseen scenes in the real world.

\subsection{Clustering Instance Mix}
Considering the irregular and unordered attributes of point clouds, we propose the CINMix to generate diverse and realistic source scenes under geometry constraints. 
CINMix includes two steps: the density-based class clustering and geometry constrained mixing.

\begin{figure}
    \centering
    \includegraphics[width=.99\linewidth]{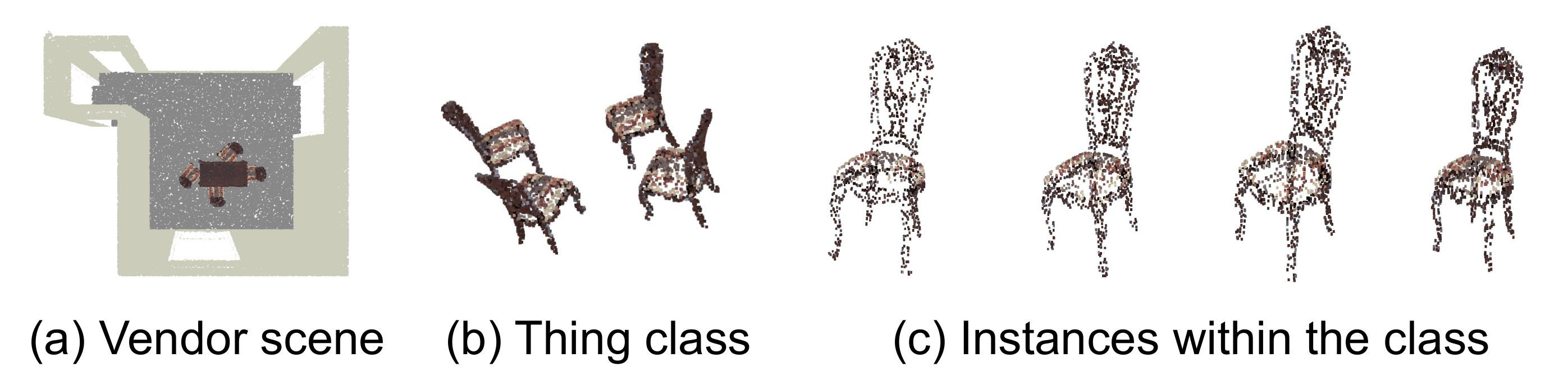}
    \caption{Visualization of one thing class (\ie, chair) and its clustered instance groups after density-based class clustering.}
    \label{fig:cinmix-clustering}
\end{figure}

\begin{figure}
    \centering
    \includegraphics[width=.99\linewidth]{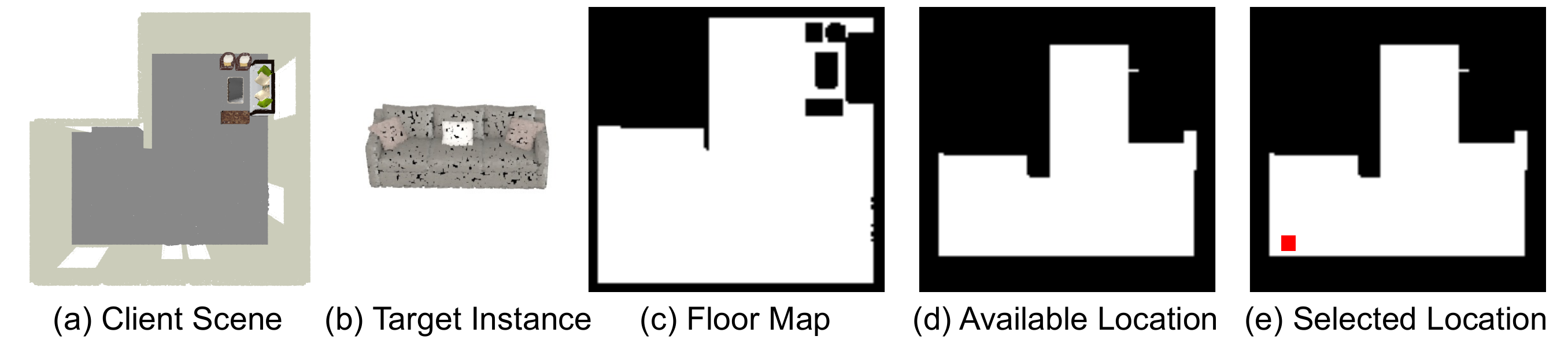}
    \caption{Procedures of geometry constrained mixing.}
    \label{fig:cinmix-mix}
\end{figure}

\noindent\textbf{Density-based Class Clustering.}
{As can be seen from Fig.~\ref{fig:cinmix-clustering}~(a) and (b), 2D based ClassMix~\cite{olsson2021classmix} cannot be used since one class in the 3D space may not be constrained in a fixed size range like that in the 2D images.}
For example, the ``sofa'' class in a living room cannot be directly put into another kid room, since the living rooms are commonly larger and the sofas can spread across the room.
Instead, the thing class, \eg, chair and table, can be easily separated into different parts by density-based clustering. The clustering parts can contain one instance or multiple nearby instances, but each part is within a relatively small scale, which can be mixed into other scenes. 
Motivated by this, we use DBSCAN~\cite{dbscan} for each thing class in the vendor scene to get one or multiple instances within a class. The clustering results are shown in Fig.~\ref{fig:cinmix-clustering}(c).

\noindent\textbf{Geometry Constrained Mixing.}
Given the clustering instances, where to fuse them into the client scene remains a problem. Since the point clouds of vendor and client scenes are not aligned, directly concatenating them may lead to unrealistic results. 
To this end, we introduce two major geometry constraints for the mixing: the mixed instances should be (1) on the floor, and (2) have no overlap with the existing classes in the client scene. The overall mixing procedures are shown in Fig.~\ref{fig:cinmix-mix}. First, a non-class floor map is obtained from the client scene. 
Then, erosion is applied to the floor map by treating the shape of the clustering instance as the kernel size.
Finally, the clustering instance is inserted into one of the available locations randomly. 
Note that we mix multiple clustering instances to one client scene and apply random rotation along  $z$ axis for each instance to improve the diversity of our CINMix augmentation.

\subsection{Non-parametric Multi-prototypes}
Different point patterns, represented by noise and occlusion, is another severe domain shift type limiting the generalization ability. We use virtual scan simulation~\cite{ding2022doda} to add noise and occlusion into the synthetic scenes, which can augment the point patterns of these scenes. 
However, the addition of the noise and occlusion leads the synthetic data to have a larger intra-class variance that is similar to the variance of the complex real-world data (see Fig.~\ref{fig:intro}).
To tackle the enlarged intra-class variance,
we introduce the non-parametric multi-prototypes.
These multi-prototypes can encourage the model to discover intra-class discriminative patterns and optimize the mining of the instance-specific information. 

\noindent\textbf{Multi-prototypes Initialization.}
The multi-prototypes are initialized by clustering the features of source data. Specifically, we first calculate the mean feature of each class $c$ for each augmented source sample $x^i$: %containing $n$ points:
\begin{equation}
    \Bar{f}_c^i = \frac{\sum_{j=1}^n f(x^i_j) * \mathbbm{1}\left(y^i_j==c \right)}{\sum_{j=1}^n \mathbbm{1}\left(y^i_j==c \right)},
\end{equation}
where $n$ denotes the number of points in $x^i$.
Given all the class-wise features in the source domain, we cluster features of each class into $K$ clusters and get the $K$ cluster centroids as the multi-prototypes $P_c \in \mathbb{R}^{K\times D}$. K-means++~\cite{kpp} is adopted in this paper.
\begin{equation}
    P_c = \operatorname{KPP}([\Bar{f}_c^1, \Bar{f}_c^2, \cdots, \Bar{f}_c^n], K),
\end{equation}
where $\operatorname{KPP}$ denotes the K-means++ clustering algorithm and $K$ is the number of clusters.

\noindent\textbf{Momentum Update via Optimal Transport.}
Since feature representations are changing along with the training, the initialized multi-prototypes should be updated for accurate representations. 
Intuitively, each prototype in multi-prototypes can be updated by the feature(s) that are closest to the prototype. However, this might lead to a degenerate solution that only one prototype is updated when all features are close to this specific prototype. 
Inspired by \cite{ym2020self,zhou2022rethinking}, we adopt the solution of optimal transport to split the features uniformly to $K$ prototypes. Specifically, $n_c$ features of class $c$ in the mini-batch $X_c \in \mathbb{R}^{n_c \times D}$ is expected to be assigned to $K$ prototypes $P_c \in \mathbb{R}^{K\times D}$ uniformly, and each feature can only be assigned to one prototype. Thus, the assignment matrix, denoted as $Q_c \in \mathbb{R}^{n_c \times K}$, should satisfy three constraints:
\begin{equation}
\label{eq:opt}
    Q_c \in \{0, 1\}^{n_c \times K}, Q_c \cdot \mathbf{1}^K = \mathbf{1}^{n_c}, Q_c^\top \cdot \mathbf{1}^{n_c} = \frac{n_c}{K} (\mathbf{1}^K)^\top,
\end{equation}
where $\mathbf{1}^{n_c}$ and $\mathbf{1}^K$ denote the vectors of all ones of $n_c$ and $K$ dimensions, respectively.
$Q$ under the constraint of Eq.~\ref{eq:opt} can be solved from the perspective of optimal transport~\cite{ym2020self,cuturi2013sinkhorn}. We adopt the \text{Sinkhorn-Knopp algorithm}~\cite{cuturi2013sinkhorn}, which relaxes the constraints by introducing a regularization term:
\begin{equation}
\label{eq:sk}
\max_{Q_c}  {\langle Q_c, X_c P_c^\top \rangle_{\mathrm{F}} + \frac{1}{\lambda} h(Q_c) }, 
\end{equation}
where $\langle \cdot \rangle_{\mathrm{F}}$ denotes the Frobenius inner product between two matrices, and $h$ denotes the entropy. 
The maximizer of Eq.~\ref{eq:sk} can be written as:
\begin{equation}
    Q_c = \text{diag}(\alpha) \exp(\lambda X_c P_c^\top) \text{diag}(\beta),
\end{equation}
where $\alpha \in \mathbb{R}^{n_c}$ and $\beta \in \mathbb{R}^K$ are scaling vectors obtained via several steps of \text{Sinkhorn-Knopp iteration}~\cite{cuturi2013sinkhorn}, and $\lambda$ is a parameter that controls the smoothness of distribution. During training, $Q_c$ is obtained at each iteration for each class, and then the multi-prototypes are updated by the point cloud features based on $Q_c$:
\begin{equation}
    P_c \leftarrow m \cdot P_c + (1-m) \cdot \frac{K}{n_c} Q_c^\top X_c,
\end{equation}
where $m$ is the update momentum.

\noindent\textbf{Prototypical Similarity Loss.}
Given the multi-prototypes $P_c \in \mathbb{R}^{K\times D}$ for class $c$, we define the probability of one point $x_j$ to the class $c$ as the maximum similarity to the multi-prototypes:
\begin{equation}
    s(x_j,c) = \max \left( f(x_j) \cdot P_c^\top \right).
\end{equation}

Instead of the global representation for each class, $s(x_j,c)$ can model the most similar representation of the same class and the most confusing representation from other classes. Optimizing over such probability can force the model to discover intra-class discriminative patterns and support the learning of instance-specific details. Consequently, we define the prototypical similarity loss as:
\begin{equation}
    \mathcal{L}_{\text{proto}}(x, y) = - \frac{1}{n} \sum_{j=1}^n y_j \log{\frac{ \exp{ \left( s(x_j, c_j) \right) }} {\sum_{k = 1}^{N_C} \exp{ \left( s(x_j, c_k) \right) } }},
\end{equation}
where $n$ and $N_C$ denote the number of point clouds in $x$ and the number of classes, respectively.

\noindent\textbf{Prediction Rectification.}
Compared with the global classifier, the non-parametric multi-prototypes contain more intra-class discriminative patterns,
which can serve as the rectification for the model prediction to alleviate the instance-specific impact.
For example, armchair may be closer to the unified representation of sofa instead of chair, which may lead to an incorrect prediction from the global classifier. However, one of the multi-prototypes of the chair can represent the armchair, and integrating such information can rectify the result.
Thus, we leverage the similarity to the multi-prototypes as the weight to rectify the global classifier prediction. The weight is obtained by:
\begin{equation}
    w(x_j,c) = \frac{\exp{\left( s\left(x_j, c \right) \right)} } { \sum_{k = 1}^{N_C} \exp{\left( s\left(x_j, c_k \right) \right)}}.
\end{equation}

Then the model prediction of the global classifier $\Phi$ is rectified by $w  (x_j, c)$:
\begin{equation}
    p (x_j,c) = w  (x_j, c) * \Phi \left( f(x_j), c \right),
\end{equation}
where $p(x_j,c)$ is the rectified prediction for the point $x_j$.

\subsection{Training Objective}
Equipped with the clustering instance mix and multi-prototypes, the model is optimized by the combination of the cross entropy loss over the global classifier and the proposed prototypical similarity loss:
\begin{equation}
    \mathcal{L} = \mathcal{L}_{\text{CE}}(x, y) + \mathcal{L}_{\text{proto}}(x, y).
\end{equation}

\section{Experiments}

\begin{table*}[ht]
    \centering
        \begin{tabular}{c|l|cccccccc|c}
            \toprule
            % \multirow{1}{*}{3D-FRONT $\rightarrow$ }  %\\& \multicolumn{9}{c}{ScanNet} \\
            Target & Method & wall & floor & chair & sofa & table & door & wind. & bksf. & mIoU \\
            \midrule
            \multirow{3}{*}{\rotatebox{90}{\textbf{ScanNet}}} & Source Only & {70.24} & 86.41 & 61.53 & 31.90 & 50.95 & 6.60 & 3.02 & 31.93 & 42.82 \\
            & VSS~\cite{ding2022doda} & 68.80 & \textbf{89.05} & 57.91 & \textbf{45.49} & 47.22 & \textbf{8.47} & 10.09 & 36.42 & 45.43 \\
            & \textbf{Ours}  & \textbf{73.42} & 88.07 & \textbf{62.72} & {43.56} & \textbf{52.14} & {7.71} & \textbf{14.20} & \textbf{44.44} & \textbf{48.28} \\
            \midrule
            \multirow{3}{*}{\rotatebox{90}{\textbf{S3DIS}}} & Source Only & 68.94 & 92.63 & 50.86 & 16.61 & 47.50 & 9.00 & 0.87 & 22.74 & 38.64 \\
            & VSS~\cite{ding2022doda} & 69.71 & 94.52 & 58.11 & \textbf{26.43} & 40.83 & \textbf{21.16} & 23.14 & 49.24 & 47.89 \\
            & \textbf{Ours}  & \textbf{76.00} & \textbf{94.66} & \textbf{66.22} & 17.79 & \textbf{53.20} & 21.12 & \textbf{29.84} & \textbf{51.14} & \textbf{51.25} \\
            \bottomrule
        \end{tabular}
        \vspace{-.1in}
        \caption{Domain generalization results on 3D-FRONT $\rightarrow$ ScanNet \& S3DIS benchmark in terms of mIoU (\%).}
    \label{tab:main}
\end{table*}

\subsection{Experimental Setup}

\noindent\textbf{Datasets.}
We conduct experiments on the synthetic-to-real domain generalization setting. The synthetic dataset 3D-FRONT~\cite{3dfront} is leveraged as the training data, and the model is evaluated on two real-world datasets, \ie ScanNet~\cite{scannet} and S3DIS~\cite{s3dis}.
3D-FRONT~\cite{3dfront} is a large-scale dataset of 3D indoor scenes, containing 18,968 rooms with 13,151 3D furniture objects. 
There are 31 different scene categories and 34 semantic classes in 3D-FRONT,
and the layouts of the rooms are designed by professional experts.
Following \cite{ding2022doda}, we use 4,995 rooms as the training data and another 500 rooms as the validation.
For the real-world target domains, ScanNet~\cite{scannet} is a large-scale real-world dataset for point cloud scene understanding, containing 1,201 training, 312 validation, and 100 testing scans.
S3DIS~\cite{s3dis} is a real-world 3D semantic segmentation dataset with 271 scenes from 6 areas. Following \cite{qi2017pointnet++,ding2022doda}, 68 samples in the fifth area are used as the validation data, while the remaining 203 scenes are treated as the training data.

\noindent\textbf{Evaluation Metric.}
We use eight shared categories across the three datasets for training and evaluation, and we adopt the mean Intersection-over-Union (mIoU) over the eight categories to evaluate the model performance.

\noindent\textbf{Implementation Details.}
Following \cite{ding2022doda}, we adopt sparse-convolution-based U-Net model~\cite{sparseconv} as our backbone.
We use AdamW~\cite{adamw} optimizer with an initial learning rate 6e-4 and weight decay 0.01 to optimize the model. The polynomial decay~\cite{polydecay} with a power of 0.9 is used as the learning rate scheduler. All models are trained for 100 epochs with a batch size of 32. The number of multi-prototypes $K$, the smoothness parameter $\lambda$, and the update momentum $m$ are set to 3, 20, and 0.999, respectively. For the DBSCAN algorithm in density-based class clustering, we set the maximum distance between neighbor points to 0.2 and the number of points in a neighborhood to 100.

\noindent\textbf{Baselines.}
Previous works~\cite{zhao2022shade,robustnet,zhong2022adversarial} on DG-Seg in 2D images consider the domain shift as image style change. Hence, these works cannot be applied to 3D point clouds.
As we are the first work focusing on domain generalized semantic segmentation in 3D indoor scenes, we compare our method with the model trained only with cross entropy loss on the source data (source only). In addition, since virtual scan simulation (VSS)~\cite{ding2022doda} plays an important role in narrowing the domain gap by adding noise and occlusion, we use the source only model with VSS as a strong baseline.

\subsection{Synthetic-to-Real Domain Generalization}
In Tab.~\ref{tab:main}, we compare our model with two baselines on the 3D-FRONT $\rightarrow$ ScanNet \& S3DIS benchmark.
We make the following observations. \textbf{First}, on the ScanNet dataset, our model outperforms the source only model all the categories and achieves the mIoU of 48.28\%, yielding an improvement of 5.46\% and 2.85\% over the source only and VSS models respectively.
\textbf{Second}, our model achieves the best performance on 3 thing classes (chair, sofa, table, and bookshelf) on the ScanNet dataset, demonstrating the effectiveness of mixing the instances of thing classes and mining the intra-class variance.
\textbf{Third}, our model also achieves the state-of-the-art performance on S3DIS dataset. Without accessing any real-world data, our model obtains a mIoU of 51.25\%, outperforming the source only baseline by 12.61\%.

\begin{table*}[t]
\begin{center}
%
% \footnotesize
\setlength{\tabcolsep}{3pt}
\begin{tabular}{c|p{1.5cm}<\centering|p{1.5cm}<\centering|p{1.5cm}<\centering|p{1.5cm}<\centering|p{1.5cm}<\centering|p{1.5cm}<\centering|p{1.5cm}<\centering}
\toprule
No. & CINMix & Para. & N-Para. T & N-Para. I & ScanNet & S3DIS & Mean \\
\midrule
1 &\xmark & \textcolor{dark-green}{\cmark} & \xmark & \xmark & 45.43 & 47.89 & 46.66 \\
\midrule
2 &\textcolor{dark-green}{\cmark} & \textcolor{dark-green}{\cmark} & \xmark& \xmark & 46.23 & 48.92 & 47.58 \\
3 &\xmark & \textcolor{dark-green}{\cmark} & \textcolor{dark-green}{\cmark} & \xmark & 46.63 & 48.29 & 47.46 \\
4 &\xmark & \textcolor{dark-green}{\cmark} & \textcolor{dark-green}{\cmark} & \textcolor{dark-green}{\cmark} & 46.78 & 49.06  & 47.92 \\
5 &\textcolor{dark-green}{\cmark} & \xmark & \textcolor{dark-green}{\cmark} & \xmark & 46.88 & 46.26 & 46.57 \\
6 &\textcolor{dark-green}{\cmark} & \textcolor{dark-green}{\cmark} & \textcolor{dark-green}{\cmark} & \xmark & 48.00 & 50.02 & 49.01 \\
7 &\textcolor{dark-green}{\cmark} & \textcolor{dark-green}{\cmark} & \textcolor{dark-green}{\cmark} &\textcolor{dark-green}{\cmark} & \textbf{48.28} & \textbf{51.25} & \textbf{49.77} \\
\bottomrule
\end{tabular}
\vspace{-.1in}
\caption{Ablation studies on CINMix and non-parametric multi-prototypes. ``Para.'' denotes the parametric classifier. ``N-Para. T'' and ``N-Para. I'' denote using the non-parametric multi-prototypes in training and inference stages, respectively.}
\label{table:ablation-loss}
\end{center}
\end{table*}

\subsection{Ablation Studies}

In this section, we analyze the effectiveness of clustering instance mix (CINMix) and non-parametric multi-prototypes in our model and compare the non-parametric classifier with the parametric classifier. 

\noindent\textbf{CINMix.} 
CINMix is proposed to diversify the layouts in the source domain by extracting instances from the vendor scene and mixing them to the client scene under the two geometry constraints. The instances are obtained from the density-based clustering on the points of thing classes. 
As shown in the 2nd row of Tab.~\ref{table:ablation-loss}, CINMix can improve the performance on ScanNet and S3DIS datasets by 0.8\% and 1.03\%, respectively, which demonstrates the effectiveness of CINMix in generating diverse and realistic scenes.

\noindent\textbf{Non-parametric Multi-prototypes.}
We propose non-parametric multi-prototypes to address the intra-class variance in the 3D indoor scenes.
The proposed multi-prototypes are used in both training and inference stages to learn instance-specific details and rectify the model prediction, respectively.
\textbf{First}, comparing the 3rd row with the baseline model, introducing non-parametric multi-prototypes in the training stage can improve the performance on both of the target datasets, yielding an improvement of 0.8\% on the average mIoU.
\textbf{Second}, we compare using the parametric and non-parametric classifiers separately in the 2nd and 5th rows of Tab.~\ref{table:ablation-loss}. The non-parametric classifier can improve the performance on ScanNet but achieves worse performance than the parametric one on S3DIS. 
We conjecture that the test set of S3DIS is less complex than that of ScanNet, and thus the target data may be biased to some of the prototypes without the guidance of global representation, leading to incorrect predictions.
\textbf{Third}, when combining the parametric and non-parametric classifiers, the model can gain improvement either with or without the help of CINMix by 1.26\% and 2.19\% in the average mIoU. Note that CINMix generates more diverse scenes, which can 
boost the capacity of the non-parametric classifier.
\textbf{Finally}, the non-parametric multi-prototypes can also be used in the inference stage to rectify the model prediction by alleviating the impact of instance-specific information. As shown in the last row of Tab.~\ref{table:ablation-loss}, the rectification can slightly improve the performance on ScanNet by 0.28\% in mIoU but gains great improvement on S3DIS by 1.23\% in mIoU. The reason is that with the guidance of global representation, the multi-prototypes can better address instance-specific details of S3DIS instead of biased to some specific prototypes. 
All the results demonstrate the effectiveness of non-parametric multi-prototypes in ameliorating the intra-class variance and improving the generalization ability.

\begin{table}[t]
\begin{center}
%
% \footnotesize
\setlength{\tabcolsep}{3pt}
\begin{tabular}{l|c|c|c}
\toprule
Augmentation & ScanNet & S3DIS & Mean \\
\midrule
w.o. Aug & 46.78 & 49.06 & 47.92 \\
\midrule
Mix3D~\cite{nekrasov2021mix3d} & 48.07 & 45.36 & 46.70 \\
Cuboid Mixing~\cite{ding2022doda}  & 47.74 & 45.58 & 46.66 \\
\textbf{CINMix} & \textbf{48.28} & \textbf{51.25} & \textbf{49.77} \\
\bottomrule
\end{tabular}
\vspace{-.1in}
\caption{Comparison with different augmentation techniques.}
\label{table:augmentation}
\end{center}
\end{table}

\subsection{Evaluation}

\noindent\textbf{Data Augmentation Techniques.}
To verify the effectiveness of CINMix, we conduct experiments on different data augmentation techniques, including Mix3D~\cite{nekrasov2021mix3d} and Cuboid Mixing~\cite{ding2022doda}. Mix3D directly concatenates two scenes after translating them to the origin.
Cuboid Mixing first partitions two scenes into several cuboids with varying sizes and then randomly mixes them with spatial permutation. 
In Tab.~\ref{table:augmentation}, compared with the baseline model without additional augmentation, all three augmentation techniques can improve the performance on ScanNet. However, Mix3D and Cuboid Mixing deteriorate the performance on S3DIS, since the complete scenes in S3DIS suffer from the complex context introduced by these two augmentations. 
Compared with the above two methods, our CINMix can generate diverse and realistic source samples, improving the performance on both datasets by a large margin.

\begin{figure}[t]
    \centering
    \includegraphics[width=.99\linewidth]{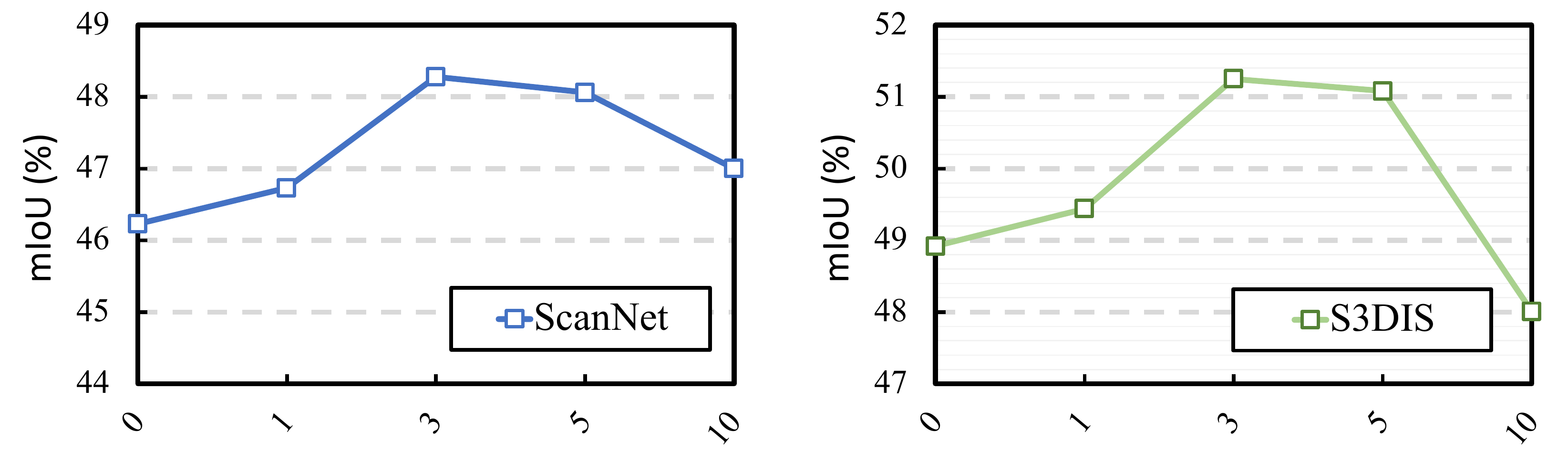}
    \vspace{-.1in}
    \caption{Parameter analysis on $K$.}
    \label{fig:param}
\end{figure}

\noindent\textbf{Number of Multi-prototypes.}
The number of multi-prototypes $K$ is an important hyper-parameter in our model, which can affect the generalization ability in both training and inference stages. We analyze $K$ in Fig.~\ref{fig:param}. \textbf{First}, introducing an additional non-parametric classifier $K=1$ can improve the performance by compensating the learned representation of the parametric classifier. \textbf{Second}, the model performance is improved along with the increase of $K$ and peaks at $K=3$. \textbf{Third}, the performance degrades when further increasing $K$, since the model cannot discover representative pattern for each prototype when $K$ is quite large.
Consequently, $K$ is set to 3 in our experiments.

\begin{table}[t]
    \centering
    \begin{tabular}{l|c|c}
    \toprule
    Method & Benchmark & mIoU \\
    \midrule
    Source Only & \multirow{2}{*}{ScanNet $\rightarrow$ S3DIS} & 58.14 \\
    \textbf{Ours} & & \textbf{62.33} \\
    \midrule
    Source Only & \multirow{2}{*}{S3DIS $\rightarrow$ ScanNet} &  43.17 \\
    \textbf{Ours} & & \textbf{45.70} \\
    \bottomrule
    \end{tabular}
    \vspace{-.1in}
    \caption{Domain generalization results on ScanNet $\rightarrow$ S3DIS and S3DIS $\rightarrow$ ScanNet benchmarks.}
    \label{tab:dg-r2r}
\end{table}

\begin{figure*}[ht]
    \centering
    \includegraphics[width=.8\linewidth]{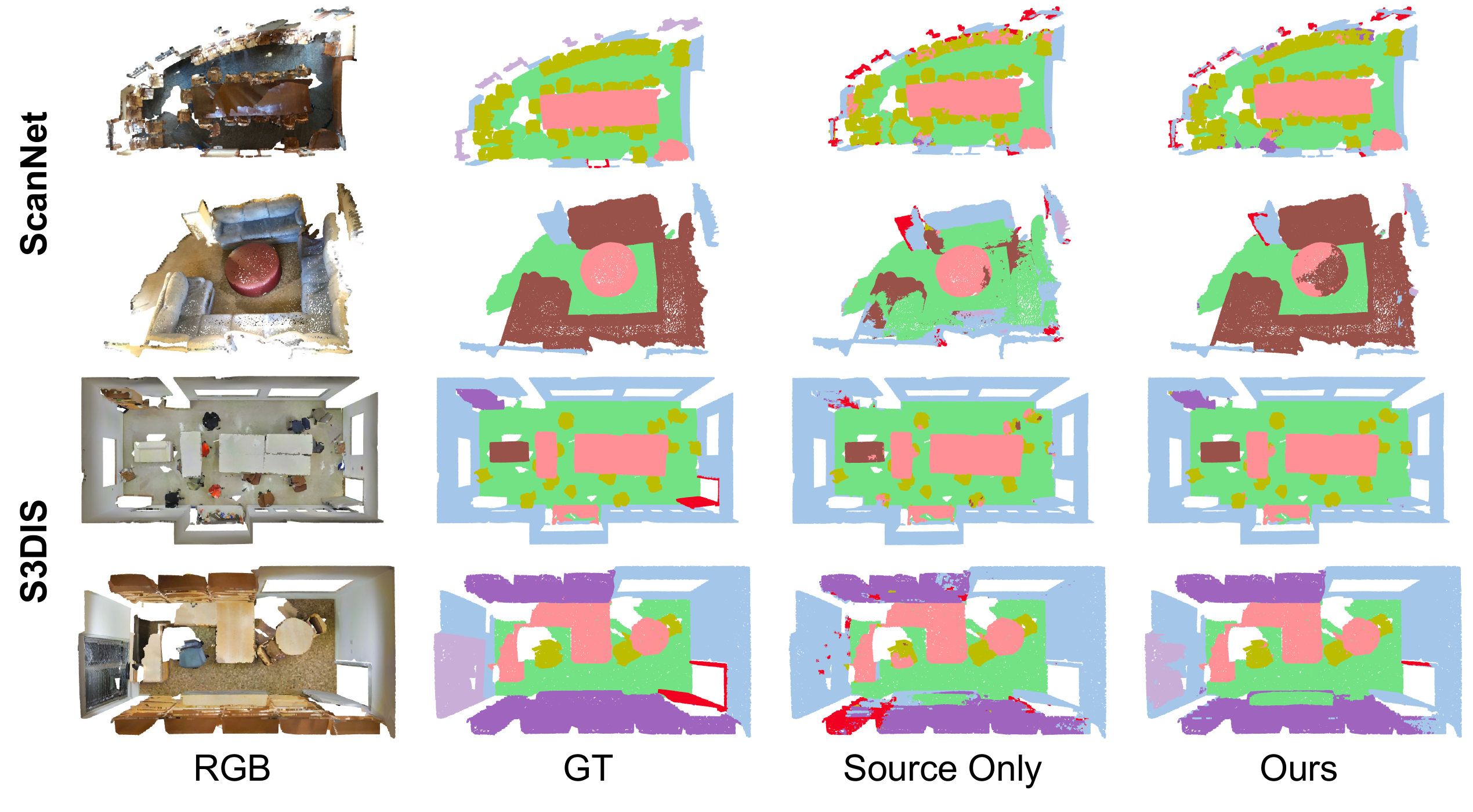}
    \vspace{-.15in}
    \caption{Qualitative comparison of segmentation results on ScanNet and S3DIS.}
    \label{fig:visual-pred}
\end{figure*}

\begin{figure}[ht]
    \centering
    \includegraphics[width=.99\linewidth]{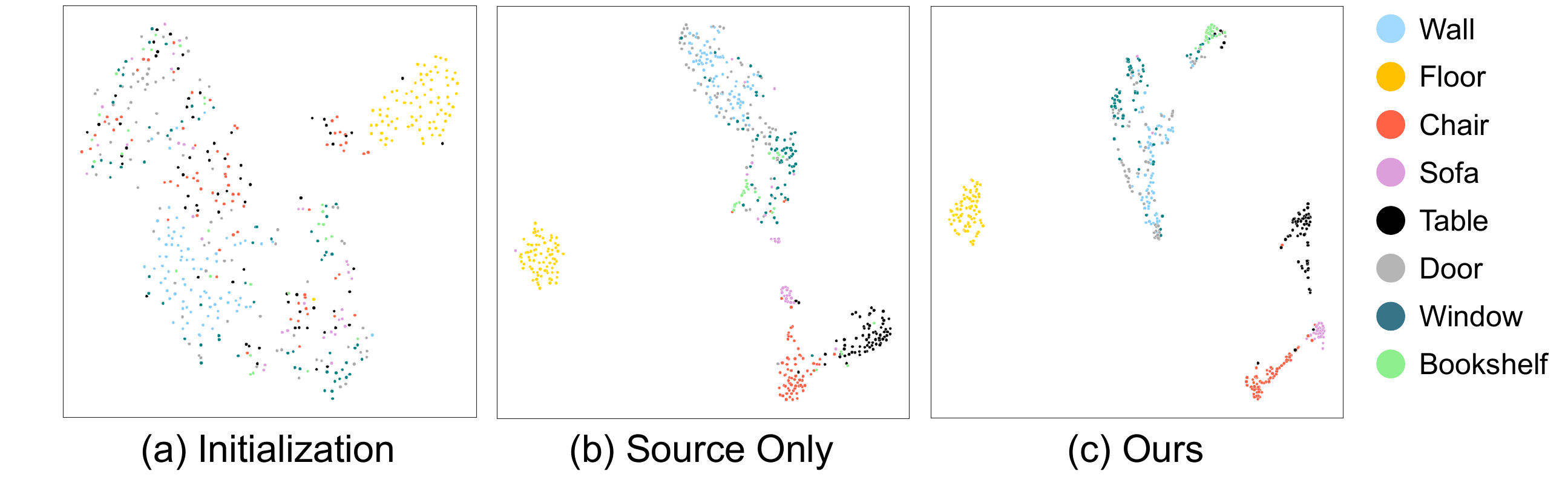}
    \vspace{-.15in}
    \caption{t-SNE visualization of feature distrib. on ScanNet val.}
    \label{fig:visual-tsne}
\end{figure}

\subsection{Real-to-Real Domain Generalization}

Despite focusing on synthetic-to-real domain generalization in this paper, we conduct experiments on the real-to-real domain generalization setting to further demonstrate the effectiveness of our method. 
Specifically, we use two benchmarks, ScanNet $\rightarrow$ S3DIS and S3DIS $\rightarrow$ ScanNet, and the eight shared classes are used to train and evaluate the model. 
VSS data augmentation is not adopted since the real-world training dataset has already contained noise and occlusion. As shown in Tab.~\ref{tab:dg-r2r}, our model outperforms the source only baseline by 4.19\% and 2.53\% in mIoU on ScanNet $\rightarrow$ S3DIS and S3DIS $\rightarrow$ ScanNet benchmarks, respectively. Such results demonstrate the versatility of the proposed CINMix and non-parametric multi-prototypes.

\subsection{Visualization}

\noindent\textbf{Comparison of Qualitative Results.}
We compare the segmentation results among the ground truth, baseline (\ie, source only), and our model. We obtain two observations from Fig.~\ref{fig:visual-pred}. \textbf{First}, our model consistently outperforms the baseline on different indoor scenes (\eg, office and living room). \textbf{Second}, our model can better deal with both hard (door and bookshelf) and similar (chair, table, and sofa) classes.
The above two observations demonstrate that our model has a strong generalization ability to segment point clouds in different unseen indoor scenes.

\noindent\textbf{Visualization of the Target Distributions.}
To better understand the effectiveness of our method, we visualize the distribution of ScanNet validation data in Fig.~\ref{fig:visual-tsne}.
The visualized features are the mean feature of each class in each sample, of which the dimension is reduced by t-SNE~\cite{tsne}.
As illustrated in Fig.~\ref{fig:visual-tsne}, compared with the baseline, our model pushes the features of the same class more compact and pulls the features of different classes more discriminating. Interestingly, the floor class can be easily separated even without training, but all the models cannot well address the confusion of door, window, and wall. Such observations suggest that CINMix and multi-prototypes lead the model to learn more generalizable representations that can perform well on unseen domains, but the segmentation of hard classes is still far from satisfactory, which deserves further investigation by future works.

\section{Conclusion}

In this paper, we introduce a new setting of domain generalized semantic segmentation in 3D indoor scenes, which trains a robust model with synthetic data but aims at segmenting real-world data of unseen domains. 
By investigating this challenging but practical problem, we find that the domain gap between synthetic and real-world data mainly lies in different layouts and point patterns. Consequently, we propose the clustering instance mix augmentation to diversify the source layouts, and the non-parametric multi-prototypes to handle the intra-class variance enlarged by augmented point patterns. 
Equipped with these designs, our model gains significant generalization ability and outperforms the baseline models by a large margin on two real-world benchmark datasets. %

\noindent\textbf{Limitations.}
Our model achieves good performance on the real-world datasets, but the target datasets are limited, containing only several kinds of rooms, \eg, living room, bedroom and office, which cannot cover all the situations in the real-world. In addition, we only focus on the 8 shared classes in this paper, but more different categories exist in our daily life. 
The domain generalization on more diverse and complex scenes deserves investigation in the future.

%%%%%%%%% REFERENCES
{\small
\bibliographystyle{ieee_fullname}
\bibliography{egbib}
}

\end{document}